\documentclass[sigconf]{acmart}
\renewcommand\footnotetextcopyrightpermission[1]{} 
\usepackage{multirow}
\usepackage{threeparttable}
\usepackage{chngcntr}
\usepackage{graphicx}
\usepackage{fancyvrb}

\AtBeginDocument{%
  }

\begin{document}

\title{Towards Robust and Controllable Text-to-Motion via Masked Autoregressive Diffusion}

\author{Zongye Zhang}
\orcid{0009-0002-8149-0533}
\affiliation{
  \institution{State Key Laboratory of Virtual Reality Technology and Systems}
  \city{Beijing}
  \country{China}
}
\email{zhangzongye@buaa.edu.cn}

\author{Bohan Kong}
\affiliation{
  \institution{State Key Laboratory of Virtual Reality Technology and Systems}
  \city{Beijing}
  \country{China}
}
\email{bohankong@buaa.edu.cn}

\author{Qingjie Liu}
\affiliation{
  \institution{State Key Laboratory of Virtual Reality Technology and Systems}
  \city{Beijing}
  \country{China}
}
\affiliation{
  \department{Hangzhou Innovation Institute}
  \institution{Beihang University}
  \city{Hangzhou}
  \state{Zhejiang}
  \country{China}
}
\email{qingjie.liu@buaa.edu.cn}
\authornote{Corresponding author.}

\author{Yunhong Wang}
\affiliation{
  \institution{State Key Laboratory of Virtual Reality Technology and Systems}
  \city{Beijing}
  \country{China}
 }
 \affiliation{
  \department{Hangzhou Innovation Institute}
  \institution{Beihang University}
  \city{Hangzhou}
  \state{Zhejiang}
  \country{China}
}
\email{yhwang@buaa.edu.cn}

\renewcommand{\shortauthors}{Zongye Zhang, Bohan Kong, Qingjie Liu, and Yunhong Wang}
\acmConference[MM '25]{Proceedings of the 33rd ACM International Conference on Multimedia}{October 27--31, 2025}{Dublin, Ireland}

\thanks{© {Owner/Author | ACM} {2025}. This is the author's version of the work. It is posted here for your personal use. Not for redistribution. The definitive Version of Record was published in Proceedings of the 33rd ACM International Conference on Multimedia, http://dx.doi.org/10.1145/3746027.3754748.}

\begin{abstract}
Generating 3D human motion from text descriptions remains challenging due to the diverse and complex nature of human motion. While existing methods excel within the training distribution, they often struggle with out-of-distribution motions, limiting their applicability in real-world scenarios. Existing VQVAE-based methods often fail to represent novel motions faithfully using discrete tokens, which hampers their ability to generalize beyond seen data. Meanwhile, diffusion-based methods operating on continuous representations often lack fine-grained control over individual frames. To address these challenges, we propose a robust motion generation framework MoMADiff, which combines masked modeling with diffusion processes to generate motion using frame-level continuous representations. Our model supports flexible user-provided keyframe specification, enabling precise control over both spatial and temporal aspects of motion synthesis. MoMADiff demonstrates strong generalization capability on novel text-to-motion datasets with sparse keyframes as motion prompts. Extensive experiments on two held-out datasets and two standard benchmarks show that our method consistently outperforms state-of-the-art models in motion quality, instruction fidelity, and keyframe adherence. The code is available at: https://github.com/zzysteve/MoMADiff
\end{abstract}

\keywords{Human Motion Generation, Text-to-Motion, Masked Modeling, Diffusion Model}

\maketitle

\section{Introduction}

\begin{figure}[t]
  \centering
  \includegraphics[width=0.95\linewidth]{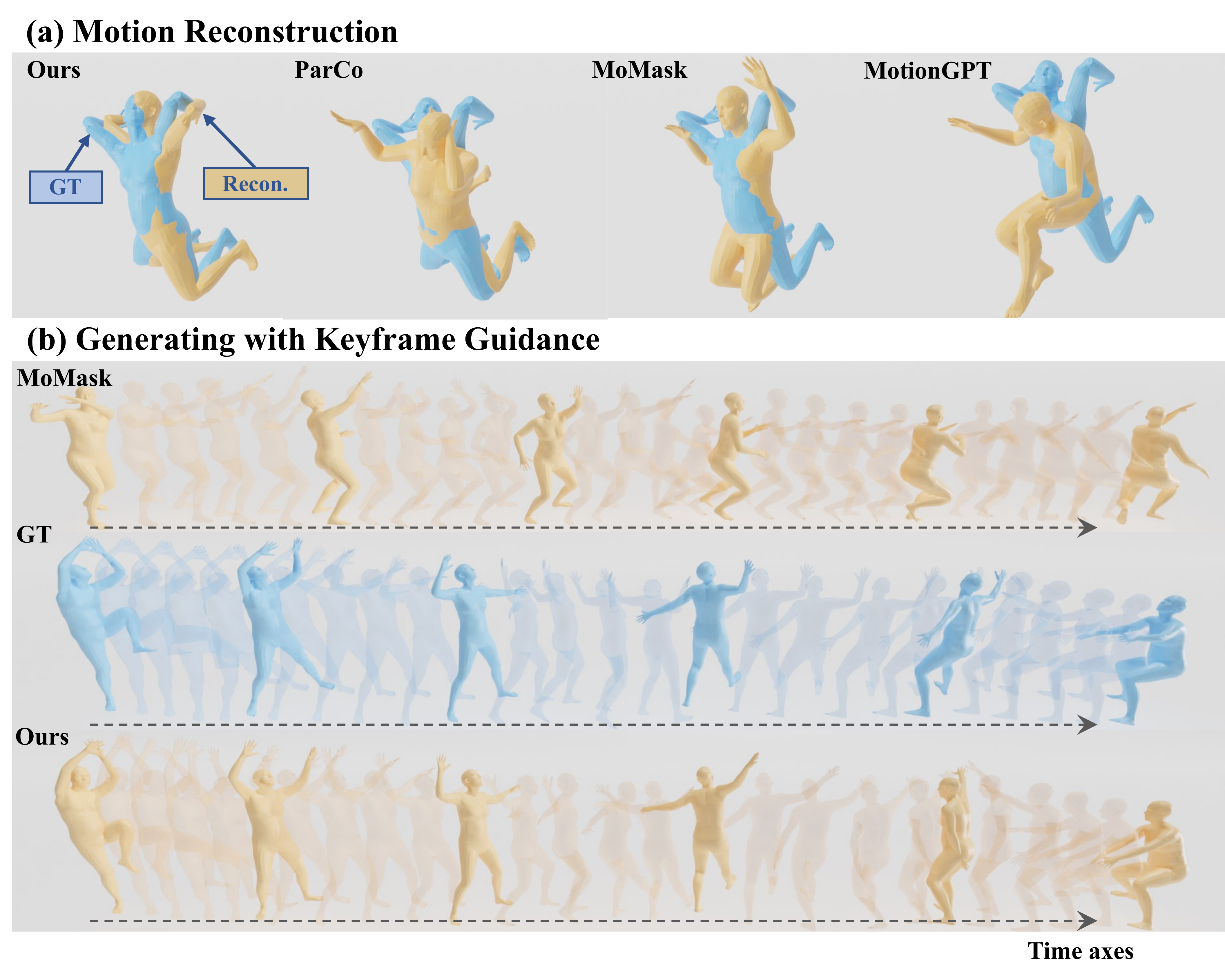}
  \caption{(a) Motion reconstruction on out-of-distribution motions using different encoders. (b) Motion generation guided by several keyframes.}
  \label{fig:vq_failure}
  \Description{}
\end{figure}

Generating 3D human motion conditioned on various inputs has received widespread attention in the past few years, with broad applications spanning virtual reality, human-machine interaction, robotics, and video games. Among these conditional modalities, text-conditioned human motion generation~\cite{ahuja2019languagepose, petrovich2022temos, guo2022generating, tevet2022motionclip, petrovich2023tmr, guo2022tm2t, jiang2023motiongpt, zhang2023generating, kong2023prioritycentric, zhong2023attt2m, zhang2024motiongpt, zhou2024avatargpt, pinyoanuntapong2024mmm, guo2024momask, pinyoanuntapong2025bamm, hosseyni2025bad, chen2024sato, zhang2024motiondiffuse, zhang2023remodiffuse, tevet2022human, kim2023flame, chen2023executing, zhou2025emdm, sampieri2025lengthaware, huang2024stablemofusion} has been at the forefront of research due to the inherent user-friendliness of natural language. However, accurately generating human motions that closely align with text descriptions remains challenging due to the highly diverse and complex nature of human motion.

Existing methods~\cite{guo2022tm2t, jiang2023motiongpt, zhang2023generating, kong2023prioritycentric, zhong2023attt2m, zhang2024motiongpt, zhou2024avatargpt, pinyoanuntapong2024mmm, guo2024momask, pinyoanuntapong2025bamm, hosseyni2025bad} have achieved impressive results by leveraging VQ-VAE and its variants, which encode motions into discrete tokens, effectively transforming motion generation from a regression problem to a classification problem. However, due to the inherent limitations of the codebook structure, VQ-VAE tends to store existing motions rather than generalize beyond them. While these models can generate and reconstruct motions accurately within the training data distribution, they often struggle with out-of-distribution motions, leading to information loss and suboptimal motion perception, as illustrated in Figure~\ref{fig:vq_failure}. This limitation hampers their ability to maintain high-quality generation when encountering motions not present in the training set, ultimately restricting their applicability in real-world scenarios. Given that training datasets cannot comprehensively cover all possible human motions, it is common for user-intended motions to be poorly represented or even absent, as illustrated in Figure~\ref{fig:vq_failure}(a).

Previous diffusion-based methods operate directly in continuous motion spaces, such as raw motion data~\cite{zhang2024motiondiffuse, tevet2022human, yang2023synthesizing, zhang2023remodiffuse, kim2023flame} or VAE-encoded latent representations~\cite{chen2023executing, dai2024motionlcm}. This inherent property allows them to avoid the limitations of discrete token representations and supports high-quality motion generation. However, these methods primarily perform segment-level modeling, generating all frames of a motion sequence at once. This design makes it difficult to modify or adjust specific frames while preserving overall motion consistency and quality. Although recent efforts have introduced mechanisms for incorporating finer control into diffusion-based models, they are typically limited to coarse, semantic-level guidance~\cite{tevet2022human, chen2024motionclr} or trajectory-based control~\cite{dai2024motionlcm}. As a result, these methods still lack fine-grained temporal control, restricting the users from precisely defining or editing motion details during generation.

To address this challenge, we propose \textbf{MoMADiff}, a framework that integrates the strengths of continuous motion spaces into masked modeling, enabling robust motion representation while preserving high-quality generation. Specifically, we introduce a VAE that supports bidirectional transformation between motion sequences and frame-wise continuous latent representations, enabling precise and fine-grained motion reconstruction. To generate these latent motion features, we employ a lightweight MLP-based diffusion head integrated with a masked autoregressive model, building on insights from~\cite{li2024autoregressive}.

In traditional character animation, the artists typically sketch keyframes first and then produce the in-between motions. Inspired by this workflow, our model first generates keyframes corresponding to the text prompts, and then recursively infers the remaining frames to complete the motion sequence. Notably, our approach offers the flexibility to either generate keyframes autonomously or incorporate user-provided keyframes. This enables the model to synthesize novel actions beyond the training distribution, guided by several specified keyframes and text instructions, as illustrated in Figure~\ref{fig:vq_failure}(b). By combining accurate motion modeling via continuous representations with flexible spatial and temporal control, our framework supports various applications, including out-of-distribution motion synthesis, long-sequence generation, and temporal motion editing.

To evaluate the robustness of our proposed method, we conduct experiments on two held-out datasets that are not used during training, simulating real-world application scenarios. Compared to discrete token-based approaches, our model demonstrates stronger control capabilities and improved robustness. In addition, we benchmark our model on two widely adopted text-to-motion datasets to compare with existing methods. Our method achieves superior performance in terms of keyframe adherence, motion quality, and instruction fidelity. Furthermore, it consistently outperforms current diffusion-based models on standard benchmarks.

Our contributions can be summarized as follows.
\begin{itemize}
\item We propose a frame-wise motion VAE that encodes human motions into sequences of continuous tokens, enabling accurate motion reconstruction and robustness across unseen datasets.
\item We introduce a masked autoregressive diffusion model that facilitates fine-grained and controllable human motion generation based on continuous frame-level tokens.
\item Our proposed \textbf{MoMADiff} achieves competitive results on standard text-to-motion benchmarks and demonstrates strong generalization ability in keyframe-guided out-of-distribution motion generation.
\end{itemize}

\section{Related Work}

\begin{figure*}[t]
  \centering
  \includegraphics[width=\linewidth]{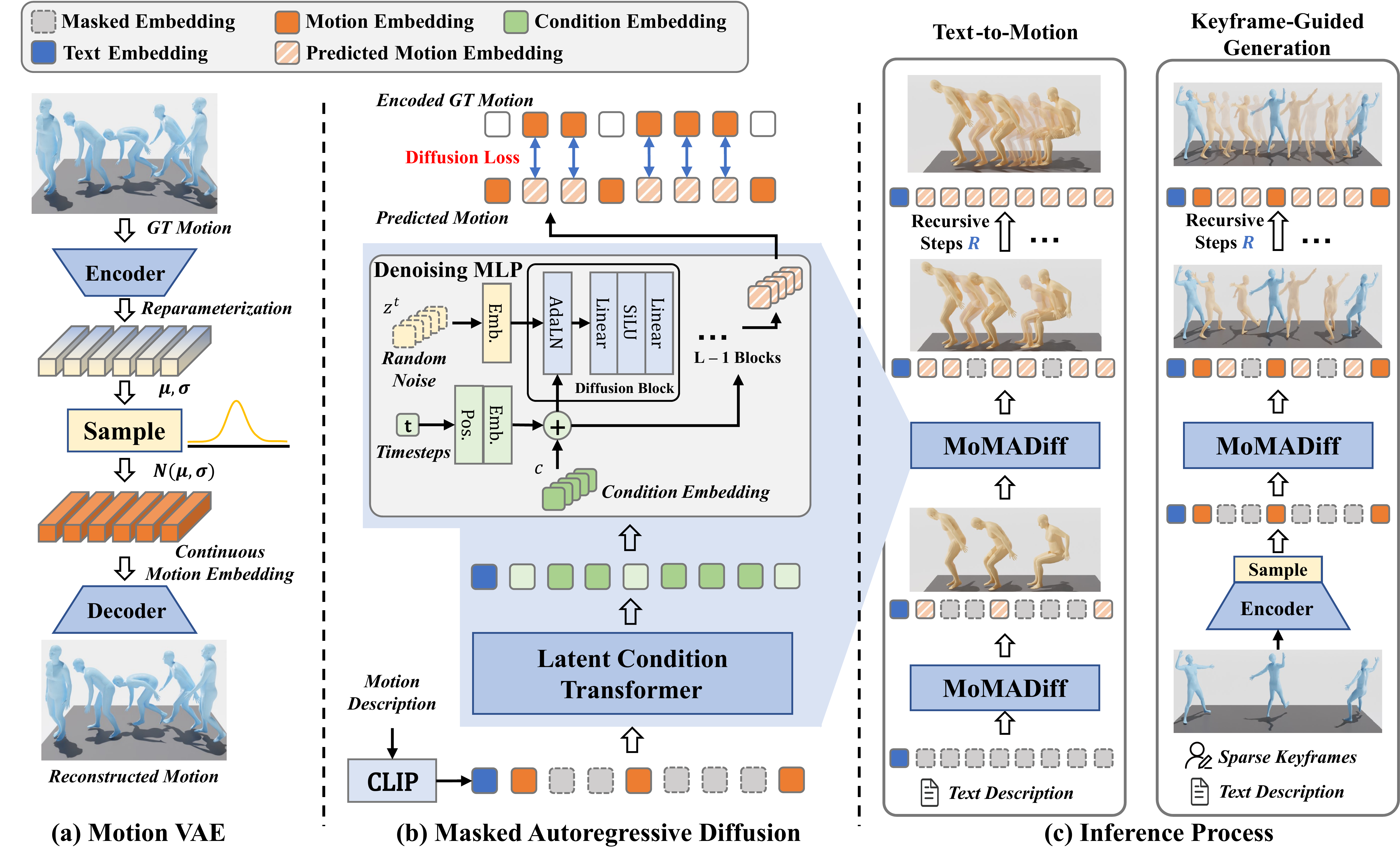}
  \caption{Overview of the proposed method. (a) A motion VAE encodes raw motion into continuous frame-wise latent embedding and decodes them for generation. (b) The autoregressive diffusion model is trained via masked modeling, using a diffusion-based prediction head to predict motion embeddings. (c) During inference, the model first generates a few keyframes and then completes the full motion sequence. It also supports sparse keyframes for controllable generation.}
  \Description{}
  \label{fig:overview}
  \vspace{-1em}
\end{figure*}

\subsection{Text-driven Human Motion Generation}

Early methods for text-driven human motion generation~\cite{ahuja2019languagepose, petrovich2022temos, guo2022generating, tevet2022motionclip, petrovich2023tmr} aim to align the distributions of motion and language features within a shared latent space using specific loss functions. These approaches typically follow a two-stage pipeline: first encoding the textual input, then decoding the corresponding motion sequence from the latent representation.

Inspired by the success of auto-regressive models in language generation, recent works~\cite{guo2022tm2t, jiang2023motiongpt, zhang2023generating, kong2023prioritycentric, zhong2023attt2m, zhang2024motiongpt, zhou2024avatargpt} have proposed autoregressive frameworks based on discrete motion representations. These models generate motion token-by-token in a sequential manner, where each token is predicted based on the previously generated ones. However, this strictly causal design limits the model's ability to capture long-range temporal dependencies and bidirectional context, which can be critical for coherent and complex motion synthesis.

To overcome this limitation, BERT-style masked modeling methods~\cite{pinyoanuntapong2024mmm, guo2024momask, pinyoanuntapong2025bamm, hosseyni2025bad} have been introduced. These approaches enable bidirectional attention over motion tokens, allowing for richer contextual understanding and support for applications such as motion editing, interpolation, and inpainting.
While these methods achieve strong performance, they rely heavily on discrete motion autoencoders to map continuous motion into token sequences. This dependency introduces a potential bottleneck: if the autoencoder lacks sufficient reconstruction fidelity, the overall system performance may degrade, particularly in terms of precision and robustness. While prior work such as SATO~\cite{chen2024sato} addresses the issue of text encoding stability, we argue that motion encoding stability is equally critical for real-world applications.

Denoising diffusion models~\cite{tevet2022human, kim2023flame, zhang2023remodiffuse, sampieri2025lengthaware} have recently emerged as powerful generative tools in the motion domain, building on their success in image synthesis. While these diffusion-based methods achieve impressive motion quality, they often suffer from slow inference due to the iterative sampling process. To address this, some work approaches compress motion sequences~\cite{chen2023executing} or reduce sampling steps through GAN~\cite{zhou2025emdm}. Hybrid designs have also emerged, integrating autoregressive components into diffusion frameworks. For instance, M2DM~\cite{kong2023prioritycentric} employs a discrete autoregressive framework built on motion tokens, while AMD~\cite{han2024amd} autoregressively invokes a diffusion module to generate motion frames. However, existing methods are built on segment-level generation, which limits their application to fine-grained motion controls.

In this work, we present a fine-grained, per-frame masked autoregressive diffusion model that operates directly in continuous motion space, enabling robust and precise motion representation. Furthermore, the proposed model exhibits strong generalization capabilities across unseen motion domains, demonstrating its practical utility in real-world applications, including motion editing, motion in-between, and spatial refinement.

\subsection{Human Motion Priors}
Pioneer methods directly regress motion sequences represented in continuous space~\cite{petrovich2021actionconditioned, tevet2022motionclip, petrovich2022temos, athanasiou2022teach, lin2023being}. As human motions are high-dimensional data sequences, learning human motion priors would ease the training process for motion generation models. VPoser~\cite{pavlakos2019expressive} learns a pose auto-encoder on the AMASS motion capture dataset, which is used as a pose encoder for motion generation~\cite{lin2023being}. Some methods model the motion using transformer VAEs~\cite{petrovich2021actionconditioned, tevet2022motionclip, petrovich2022temos, athanasiou2022teach, lin2023being} to aid the motion generation.

To alleviate the difficulty of human motion generation, methods based on the auto-regressive models~\cite{guo2022tm2t, zhang2023generating, kong2023prioritycentric, zhong2023attt2m, zhang2024motiongpt, zou2024parco, jiang2023motiongpt, zhou2024avatargpt} and bi-directional masking models~\cite{guo2024momask, pinyoanuntapong2024mmm, pinyoanuntapong2025bamm, hosseyni2025bad} use discrete autoencoders like VQ-VAE~\cite{oord2017neural}, RVQ-VAE~\cite{guo2024momask}, which convert it to a classification problem. Some methods~\cite{zou2024parco, yuan2024mogents, hong2024bipo, wan2025tlcontrol} propose to separate whole-body motion according to different body parts and quantize them using VQ-VAEs into discrete representations. These methods exhibit impressive motion reconstruction within the data domain. However, discrete encoders prefer to encode the motion to the closest ones that they have previously seen, resulting in gaps between the real-world motions and those in the training set.

The diffusion-based models are inherently well-suited for generating continuous representations. Some methods operates on raw motions~\cite{tevet2022human, zhang2024motiondiffuse, kim2023flame, zhang2023remodiffuse, yang2023synthesizing}. Inspired by latent diffusion models for image generations, MLD~\cite{chen2023executing} adopts a transformer-based autoencoder~\cite{petrovich2022temos}. However, the transformer-based autoencoder compresses the entire sequence into a single latent vector, which limits the model’s ability to capture fine-grained temporal and spatial dynamics.

In this work, we introduce a frame-wise continuous motion prior that enables temporally fine-grained encoding of motion sequences while preserving spatial fidelity. This design supports high-quality motion generation, editing, and generalization across diverse action domains.

\subsection{Neural Human Motion In-betweening}
In this paper, we study a novel application scenario that aims to generate out-of-distribution motions with sparse keyframes using models pretrained on large-scale text-to-motion datasets. Human motion in-betweening~\cite{starke2023motion, chu2024realtime, oreshkin2024motion, cohan2024flexible} is a long-established research area aimed at generating smooth and realistic transitions between specified keyframes. Recent methods are effective at producing seamless interpolations but primarily focus on kinematic transitions without incorporating high-level semantic guidance from text descriptions. Consequently, they often rely on relatively dense keyframes~\cite{oreshkin2024motion, cohan2024flexible} to achieve satisfactory performance, such as providing keyframes at intervals of 1/6 or 1/2 of a second.

In contrast, our approach leverages both fundamental human motion priors and text-based guidance during training, significantly reducing the reliance on densely provided keyframes when applied to real-world applications. This alleviates the user’s burden of manually specifying detailed motions and enables the generation of more semantically rich and diverse motion sequences.

\section{Method}

Overview of our proposed framework is illustrated in Figure~\ref{fig:overview}. Given an input motion sequence, we first encode it into a sequence of \textbf{continuous} latent tokens using a motion autoencoder, as described in Section~\ref{sec:CMA}. During training, the model learns to predict masked segments of motion latent continuous tokens through diffusion modeling (Section~\ref{sec:MoMADiff}). At inference time, the model generates motion autoregressively in a set-by-set manner, starting from encoded text prompts, as detailed in Section~\ref{sec:inference}.

\subsection{Continuous Motion Autoencoder}
\label{sec:CMA}
Most existing work~\cite{guo2022tm2t, jiang2023motiongpt, zhang2023generating, kong2023prioritycentric, zhong2023attt2m, zhang2024motiongpt, zhou2024avatargpt, pinyoanuntapong2024mmm, guo2024momask, pinyoanuntapong2025bamm, hosseyni2025bad} encodes raw motion sequences into discrete latent codes using VQ-VAEs, thereby transforming the regression task into a classification problem for downstream text-to-motion generation. However, we observe that such discrete representations often generalize poorly to motions unseen during training. In this work, we address this limitation by modeling in a continuous latent space.

Unlike prior works that employ Transformers as motion VAEs~\cite{chen2023executing} and encode motion at the sequence level, we adopt a frame-level encoding strategy using a lightweight CNN-based architecture. Following~\cite{zhang2023generating}, we construct an encoder with $l$ layers of ResNet blocks and temporal-strided convolutions, but instead optimize the model entirely in the \textbf{continuous} domain. 
Formally, a motion sequence is denoted as $X=[x_1,x_2,...,x_T]$, where each frame $x_t \in \mathbb{R}^d$ is a $d$-dimensional motion representation~\cite{guo2022generating}. Our goal is to represent the motion with a sequence of continuous latent features $Z = [z_1, z_2, \ldots, z_{\lfloor T/l \rfloor}]$, where $z_t \in \mathbb{R}^c$ and $l$ is the temporal downsampling factor corresponding to the number of temporal-strided convolution layers. More model details can be found in the supplementary material.

The motion sequence is encoded by the encoder $\mathbf{F_e}$ as $(\mu, \sigma)=\mathbf{F_e}(X)$, and the encoded representations are sampled from $Z\sim \mathcal{N}(\mu, \sigma)$ using the encoded mean $\mu$ and variation $\sigma$. The motion is reconstructed by the decoder $F_d$ as $\widetilde{X}=\mathbf{F_d}(Z)$. The network is optimized by minimizing the following loss function,
\begin{equation}
    L=L_{NLL} + w_{k}L_{KL} + w_vL_v
\end{equation}
where $w_k$ and $w_v$ are two balance factor parameters for Kullback-Leibler (KL) loss, and joint velocity loss.

The Negative Log-Likelihood Loss (NLL) supervises the reconstruction quality.
Following~\cite{rombach2022highresolution}, it is formulated as
\begin{equation}
    L_{NLL} = \frac{||\widetilde{X}-X||_1}{\text{exp}(\text{log}\sigma^2)}+\text{log}\sigma^2
\end{equation}
where $\text{log}(\sigma^2)$ is a learnable parameter representing the log-variance, and $||\cdot||_1$ denotes the L1 norm. 

To align the posterior $q(z|x)$ with the standard normal distribution $p(z)$, we use the KL divergence:
\begin{align}
L_{KL} &= D_{KL}(q(z|x)||p(z))=-\frac{1}{2}\sum(1+\text{log}\sigma^2-\mu^2-\sigma^2)
\end{align}

To improve temporal smoothness and physical plausibility, we supervise joint velocities, represented as a subset $V$ of the motion representation $X$. The loss is computed as:
\begin{align}
L_{v} &= ||\widetilde{V} - V||_1=\sum_{t=1}^T|v_t-\widetilde{v_t}|
\end{align}
where $v_t$ and $\widetilde{v}_t$ denote the ground-truth and predicted velocities respectively.

During training, latent codes $Z$ are sampled using the reparameterization technique from $\mathcal{N}(\mu, \sigma)$, and subsequently decoded to compute the reconstruction loss. This stochastic sampling introduces variability that improves the decoder's robustness to slight noise. This alleviates the reliance on perfectly denoised latent embeddings from the diffusion model and improves the overall motion quality.

\begin{table*}[t]
\caption{Quantitative evaluation on two held-out datasets for keyframe-guided text-to-motion generation.}
\vspace{-0.5em}
\label{tab:heldout}
\centering
\resizebox{\textwidth}{!}{%
\begin{tabular}{clllllllll}
\toprule
\multicolumn{2}{c}{\multirow{2}{*}{Dataset}} &
\multirow{2}{*}{Methods} &
\multicolumn{1}{c}{\multirow{2}{*}{Venue}} &
\multicolumn{3}{c}{R-Precision} &
\multicolumn{1}{c}{\multirow{2}{*}{$FID$$\downarrow$}} &
\multicolumn{1}{c}{\multirow{2}{*}{MM-Dist$\downarrow$}} &
\multicolumn{1}{c}{\multirow{2}{*}{Diversity$\uparrow$}} \\ \cmidrule{5-7}
\multicolumn{2}{c}{} &
   &
  \multicolumn{1}{c}{} &
  \multicolumn{1}{c}{Top1$\uparrow$} &
  \multicolumn{1}{c}{Top2$\uparrow$} &
  \multicolumn{1}{c}{Top3$\uparrow$} &
  \multicolumn{1}{c}{} &
  \multicolumn{1}{c}{} &
  \multicolumn{1}{c}{} \\
  
\midrule
\midrule

\multicolumn{2}{c}{\multirow{5}{*}{IDEA400}} & Real & -         &0.923$^{\pm.001}$  &0.986$^{\pm.000}$  &0.996$^{\pm.000}$  &0.000$^{\pm.000}$  &1.363$^{\pm.001}$  &15.669$^{\pm.146}$  \\
\cmidrule{3-10}
\multicolumn{2}{c}{}                        & MDM \cite{tevet2022human} &ICLR2023         &0.411$^{\pm.005}$  &0.597$^{\pm.007}$  &0.705$^{\pm.006}$  &5.559$^{\pm.316}$  &6.022$^{\pm.047}$  & \textbf{14.924$^{\pm.138}$}  \\
\multicolumn{2}{c}{}                        & ParCo \cite{zou2024parco} &ECCV2024         &0.194$^{\pm.002}$  &0.330$^{\pm.002}$  &0.435$^{\pm.002}$  &15.105$^{\pm.043}$  &8.883$^{\pm.007}$  &13.237$^{\pm.152}$  \\
\multicolumn{2}{c}{}                        & MoMask \cite{guo2024momask} &CVPR2024 &0.194$^{\pm.002}$  &0.323$^{\pm.001}$  &0.424$^{\pm.002}$  &8.799$^{\pm.050}$  &9.268$^{\pm.009}$  &14.144$^{\pm.163}$  \\
\multicolumn{2}{c}{}                        & Ours & - &\textbf{0.644$^{\pm.002}$}  &\textbf{0.812$^{\pm.001}$}  & \textbf{0.886$^{\pm.001}$}  &\textbf{3.530$^{\pm.019}$}  &\textbf{3.611$^{\pm.005}$}  &14.368$^{\pm.141}$  \\

\midrule
\midrule

\multicolumn{2}{c}{\multirow{5}{*}{Kungfu}} & Real & -         &0.861$^{\pm.003}$  &0.924$^{\pm.002}$  &0.954$^{\pm.003}$  &0.000$^{\pm.000}$  &1.760$^{\pm.005}$  &13.416$^{\pm.093}$  \\
\cmidrule{3-10}
\multicolumn{2}{c}{}                        & MDM \cite{tevet2022human} &ICLR2023         &0.285$^{\pm.011}$  &0.407$^{\pm.009}$  &0.496$^{\pm.011}$  &19.218$^{\pm.453}$  &8.006$^{\pm.065}$  &9.294$^{\pm.074}$  \\
\multicolumn{2}{c}{}                        & ParCo \cite{zou2024parco} &ECCV2024         &0.079$^{\pm.005}$  &0.133$^{\pm.004}$  &0.180$^{\pm.005}$  &29.205$^{\pm.021}$  &3.347$^{\pm.008}$  &9.175$^{\pm.083}$  \\
\multicolumn{2}{c}{}                        & MoMask \cite{guo2024momask} &CVPR2024 &0.061$^{\pm.006}$  &0.109$^{\pm.007}$  &0.154$^{\pm.008}$  &19.254$^{\pm.343}$  &4.494$^{\pm.090}$  &11.070$^{\pm.043}$  \\
\multicolumn{2}{c}{}                        & Ours & - &\textbf{0.701}$^{\pm.007}$  &\textbf{0.844}$^{\pm.006}$  &\textbf{0.907}$^{\pm.006}$  &\textbf{2.981}$^{\pm.070}$  &\textbf{3.794}$^{\pm.022}$  &\textbf{11.766}$^{\pm.114}$  \\
\bottomrule
\end{tabular}
}
\end{table*}

\subsection{Training Latent Motion Transformer}
\label{sec:MoMADiff}
This section introduces the Motion Masked Autoregressive Diffusion (MoMADiff) model, which recursively generates per-frame continuous latent motion representations $Z$ in a next-batch prediction paradigm with bi-directional attention. During training, the input latent sequence is randomly masked by replacing selected tokens with continuous [MASK] tokens. The masked token is a learnable parameter jointly optimized during training. The text prompt is encoded using CLIP~\cite{radfordLearningTransferableVisual2021} and appended to the beginning of the transformer input sequence. The transformer then outputs a sequence of condition tokens, which serve as conditional inputs for a lightweight diffusion prediction head to reconstruct the masked motion tokens.

Inspired by~\cite{li2024autoregressive}, we keep the design of the diffusion head lightweight, as illustrated in Figure~\ref{fig:overview}(b). The condition tokens produced by the transformer are denoted as $c$ and used to guide the diffusion process for reconstructing the masked latent motion sequence $z^0$. At each diffusion step, the condition tokens are fused with the current denoising timestep $t$ and injected into the model via AdaLN~\cite{peebles2023scalable}. The diffusion block consists of a simple feed-forward structure: a linear layer followed by a SiLU activation and another linear layer. The overall diffusion head comprises $L$ such blocks.

During training, we follow the prior denoising diffusion work on human motion generation~\cite{ramesh2022hierarchical, tevet2022human}, which predicts the original motion latent $z^0$ from its noisy counterpart $z_t$, where $t$ is sampled from uniform distribution. Note that the spatial positions of the condition tokens are preserved throughout the diffusion process, and the predicted latent motion tokens are inserted back into their corresponding masked positions. Following~\cite{li2024autoregressive}, we jointly train the diffusion module and the latent condition transformer end-to-end using the following diffusion loss:
\begin{equation}
L = \mathbf{E}_{z^0 \sim q(z^0|c),\ t \sim U[1, T]} || z^0 - G(z^t, t, c) ||_2 
\end{equation}
where $c$ denotes the condition tokens generated by the transformer, $z^0$ is the ground-truth motion latent, and $G$ represents the diffusion head. Gradients from the loss flow through the diffusion head $G$ to the transformer with condition tokens $c$, enabling joint optimization.

\begin{table}[t]
\caption{Motion reconstruction On HumanML3D.}
\label{tab:reconstruction_humanml}
\resizebox{\columnwidth}{!}{%
\begin{threeparttable}
\begin{tabular}{@{}lccclccl@{}}
\toprule
\multirow{2}{*}{Method} & \multicolumn{3}{c}{Reconstruction}&  & \multicolumn{2}{c}{Generation}        \\ \cmidrule(lr){2-4} \cmidrule(l){6-7} 
& MPJPE $\downarrow$ & PAMPJPE$\downarrow$ & ACCL$\downarrow$ &  & FID$\downarrow$& DIV$\rightarrow$\\ \midrule
\textbf{Real}  & - & - & - &&-& $9.508{^{\pm.072}}$  \\ \midrule
VPoser-t~\cite{pavlakos2019expressive} & $75.6$ & $48.6$& $9.3$ &     & $1.430$\tnote{\dag} & $8.336$\tnote{\dag}   \\
ACTOR~\cite{petrovich2021actionconditioned} & $65.3$ & $41.0$& $7.0$ &     & $0.341$\tnote{\dag} & $9.569$\tnote{\dag}   \\
MLD~\cite{chen2023executing} & $\boldsymbol{14.7}$ & $8.9$& $5.1$ &     & $0.017$\tnote{\dag} & $9.554$\tnote{\dag}   \\
\midrule
ParCo~\cite{zou2024parco}  & $53.4$ & $38.1$& $7.3$ &     & $0.021{^{\pm.000}}$ & $9.388{^{\pm.078}}$   \\
MotionGPT~\cite{zhang2024motiongpt}     & $49.7$ & $33.2$& $7.7$ &     & $0.089{^{\pm.001}}$ & $9.653{^{\pm.070}}$ \\
MoMask~\cite{guo2024momask} & $31.3$ & $ 19.2$ & $6.3$ &  & $0.020{^{\pm.000}}$ &  $9.616{^{\pm.090}}$  \\
\midrule
Ours & $16.4$ & $ \boldsymbol{3.3}$ & $\boldsymbol{3.5}$ &  & $\boldsymbol{0.001}{^{\pm.000}}$ &  $\boldsymbol{9.481}{^{\pm.080}}$  \\
\bottomrule
\end{tabular}%
\end{threeparttable}
}
\begin{tablenotes}
\footnotesize
\item\dag~Reported in paper~\cite{chen2023executing}, no 95\% CI provided.
\item$\rightarrow$ indicates the diversity of reconstruction motions should be close to real ones.
\end{tablenotes}
\end{table}

\subsection{Inference-time Strategies}
\label{sec:inference}
During inference for the text-to-motion task, all latent motion tokens are initially set to [MASK] tokens, as illustrated in Figure~\ref{fig:overview}(c). In the first step, the model generates a small number of initial frames as keyframes. Specifically, the text is encoded and put at the head of the sequence before feeding into the transformer. The latent condition transformer then produces a set of condition tokens $c$, which are passed to the diffusion head to predict the continuous motion latent sequence. The diffusion head denoises $z_T$ through $T$ denoising steps, iteratively predicting $z_{T-1}, z_{T-2}, ..., z_0$, to recover the final motion latent $z_0$. To speed up the inference, we use the DDIM~\cite{song2021denoising} sampling technique to reduce the denoising step to $T_i$ in model variants with larger training steps.

Once the initial keyframes are generated, they are re-inserted into the input sequence. Alternatively, users can provide custom keyframes to guide the generation process toward specific motion characteristics. After the initial step, the model then recursively predicts the remaining intermediate frames over $R$ steps. To control the number of frames generated at each step, we employ a cosine scheduler that enables next-set prediction. This scheduler enables the model to generate a smaller number of highly controlled frames in the early stages and gradually produce a larger number of less critical frames in later steps.

To balance between generation quality and adherence to the text prompt, we apply classifier-free guidance (CFG) to the \textbf{latent condition transformer}. Importantly, we do \textbf{not} apply CFG to the diffusion head, as the spatial and temporal structures of the motion sequence are already well captured by the condition tokens. Allowing the diffusion head to generate motion independently often results in incoherent or structurally inconsistent outputs.

This staged inference strategy ensures better motion consistency, efficient sampling, and fine-grained control over both structure and content during generation.

\begin{table}[t]
\caption{Motion reconstruction on IDEA400 \& Kungfu.}
\label{tab:recons_ood}
\resizebox{\columnwidth}{!}{%
\begin{tabular}{@{}lccclccl@{}}
\toprule
\multirow{2}{*}{Method} & \multicolumn{3}{c}{Reconstruction}&  & \multicolumn{2}{c}{Generation}        \\ \cmidrule(lr){2-4} \cmidrule(l){6-7} 
& MPJPE $\downarrow$ & PAMPJPE$\downarrow$ & ACCL$\downarrow$ &  & FID$\downarrow$& DIV$\rightarrow$\\ \midrule \midrule
\textbf{IDEA400}  & - & - & - &&-& $15.669{^{\pm.146}}$  \\ \midrule 
ParCo~\cite{zou2024parco}  & $115.3$ & $81.4$& $9.2$ &     & $3.932{^{\pm.007}}$ & $14.311{^{\pm.126}}$   \\
MotionGPT~\cite{zhang2024motiongpt}& $102.7$ & $71.1$& $9.7$ &     & $6.514{^{\pm.017}}$ & $14.148{^{\pm.119}}$ \\
MoMask~\cite{guo2024momask} & $63.8$ & $37.9$ & $9.1$ &  & $1.688{^{\pm.005}}$ &  $14.935{^{\pm.096}}$  \\
\midrule
Ours & $\boldsymbol{37.4}$ & $ \boldsymbol{17.7}$ & $\boldsymbol{6.8}$ &  & $\boldsymbol{0.154}{^{\pm.001}}$ &  $\boldsymbol{15.491}{^{\pm.145}}$  \\
\midrule \midrule
\textbf{Kungfu}    & - & - & - &&-& $13.416{^{\pm.093}}$  \\ \midrule 
ParCo~\cite{zou2024parco}  & $133.0$ & $91.9$& $21.3$ &     & $2.271{^{\pm.034}}$ & $12.302{^{\pm.108}}$   \\
MotionGPT~\cite{zhang2024motiongpt}     & $163.3$ & $114.7$& $22.0$ &     & $9.689{^{\pm.103}}$ & $10.132{^{\pm.087}}$ \\
MoMask~\cite{guo2024momask} & $99.3$ & $58.9$ & $21.3$ &  & $1.442{^{\pm.024}}$ &  $12.797{^{\pm.099}}$  \\
\midrule
Ours & $\boldsymbol{56.5}$ & $ \boldsymbol{22.9}$ & $\boldsymbol{17.0}$ &  & $\boldsymbol{0.275}{^{\pm.004}}$ &  $\boldsymbol{13.388}{^{\pm.078}}$  \\
\bottomrule
\end{tabular}%
}
\vspace{-0.5em}
\end{table}

\begin{figure*}[t]
  \centering
  \includegraphics[width=\linewidth]{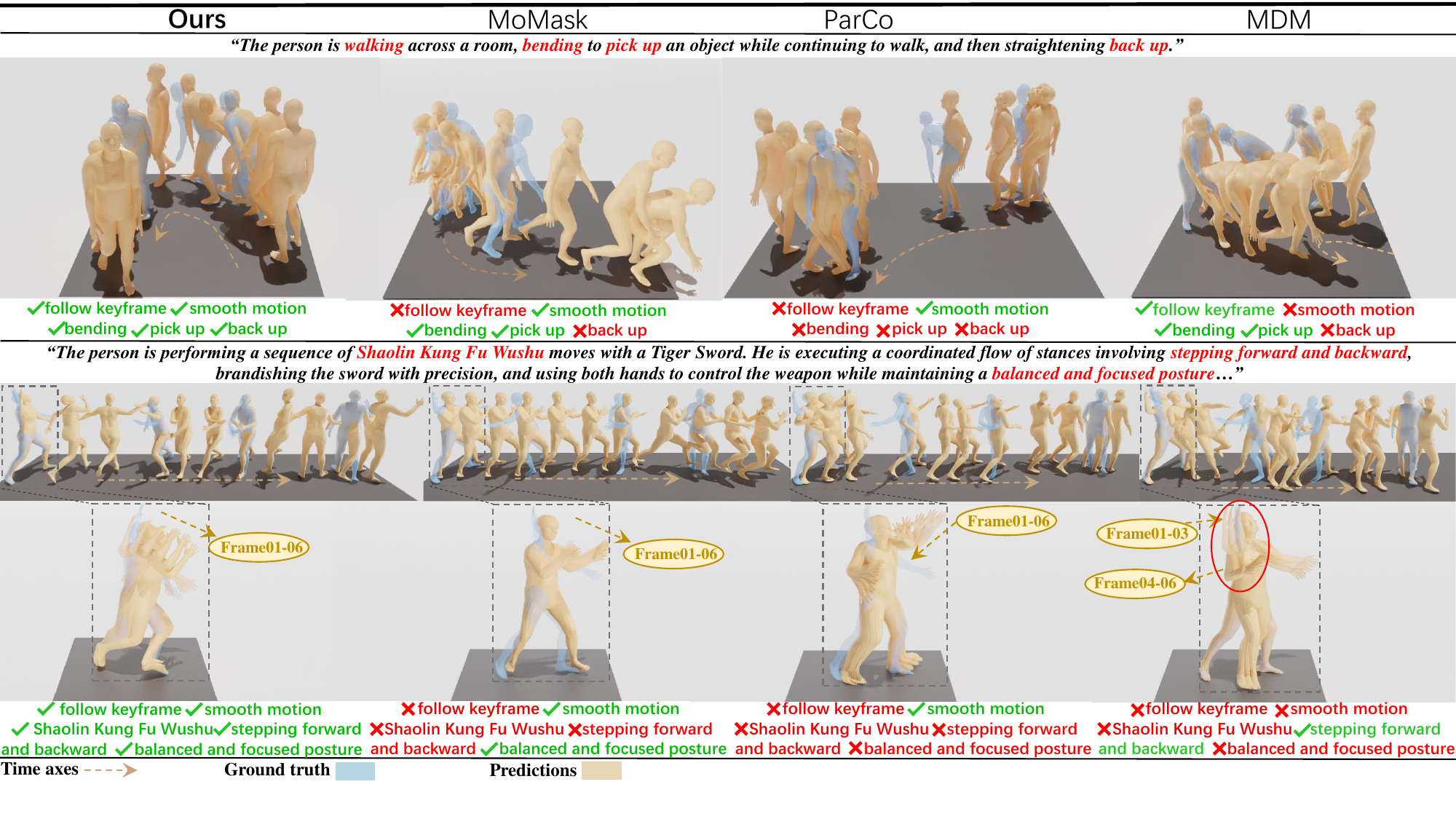}
  \vspace{-1.5em}
  \caption{Qualitative comparison of motion generation results on two held-out datasets using keyframe guidance.}
  \label{fig:qual_keyframe}
  \vspace{-0.5em}
  \Description{}
\end{figure*}

\section{Experiments}

\subsection{Datasets and Implementation Details}

\subsubsection{Datasets}

To evaluate the generalization ability of existing methods across diverse data domains, we perform cross-dataset evaluation on two subsets from Motion-X~\cite{lin2023motionx, zhang2025motion}: \textit{IDEA400} and \textit{Kungfu}. We also adopt two standard datasets for text-to-motion generation \textit{HumanML3D} and \textit{KIT-ML} to compare with current state-of-the-art models.

\textbf{HumanML3D} dataset collects the motions contains 14,616 human motions from AMASS~\cite{mahmood2019amass} dataset and HumanAct12~\cite{guo2020actionmotion}, with 44,970 textual descriptions.
\textbf{KIT-ML} dataset contains 3,911 motions from KIT~\cite{mandery2015kit} and CMU~\cite{carnegie}, and includes 6,278 descriptions. 
\textbf{IDEA400} dataset is the largest subset apart from the AMASS dataset, which contains 12,042 motion sequences with one text description each.
\textbf{Kungfu} includes 1,032 Chinese kungfu motion sequences with one semantic text description per sequence. More details can be found in the supplementary.

\subsubsection{Evaluation Metrics}

We adopt motion-text evaluator from~\cite{guo2022generating} and use the following evaluation metrics.
(1) \textit{R-Precision} measures text-motion alignment by computing Euclidean distances between a motion feature and 32 candidate text features. We report top-1, top-2, and top-3 retrieval accuracies.
(2) \textit{Frechet Inception Distance (FID)}~\cite{heusel2017gans} evaluates the distributional similarity between generated and real motions, based on features extracted by the motion encoder.
(3) \textit{Multimodal Distance (MM-Dist)} calculates the average Euclidean distance between motion features and their corresponding text features.
(4) \textit{Diversity (DIV)} assesses the variety in generated motions by computing the average Euclidean distance between 300 randomly sampled pairs.

\subsubsection{Implementation Details}

\textit{Motion VAE.} The motion VAE comprises three layers with a latent dimension of 512. It is trained using the Adam optimizer with a learning rate of 0.00005 and a batch size of 256. We employ two temporal downsampling layers, which aggregate every four consecutive frames into one latent embedding. The model is trained for 300,000 iterations, with a KL loss weighting factor of $1\mathrm{e}{-6}$ and a velocity loss weight of 0.5.

\textit{Masked Autoregressive Diffusion Model.} The transformer consists of 16 layers, each with 8 attention heads. The hidden dimension is 1024, and the output condition token dimension is 512. The diffusion head is implemented as a 4-layer MLP. We use a learning rate of 0.0001 with a linear warm-up over the first 2000 iterations. For HumanML3D, the model is trained for 600 epochs, with a decay factor of 0.1 applied at epoch 400. For the KIT-ML dataset, we train for 1500 epochs with decay at epoch 1200. To stabilize the training process, we apply an exponential moving average (EMA) to model parameters with a decay rate of 0.999. For the DDPM~\cite{ho2020denoising} variant, we use 50 diffusion steps. For the DDIM variant, we train with 1000 diffusion steps and use 100 steps during inference.

\begin{table*}[t]
\caption{Quantitative evaluation on two standard text-to-motion benchmarks.}
\label{tab:standard_bench}
\centering
\resizebox{\textwidth}{!}{%
\begin{tabular}{clllllllll}
\toprule
\multicolumn{2}{c}{\multirow{2}{*}{Dataset}} &
\multirow{2}{*}{Methods} &
\multicolumn{1}{c}{\multirow{2}{*}{Venue}} &
\multicolumn{3}{c}{R-Precision} &
\multicolumn{1}{c}{\multirow{2}{*}{$FID$$\downarrow$}} &
\multicolumn{1}{c}{\multirow{2}{*}{MM-Dist$\downarrow$}} &
\multicolumn{1}{c}{\multirow{2}{*}{Diversity$\uparrow$}}\\ \cline{5-7}
\multicolumn{2}{c}{} &
   &
  \multicolumn{1}{c}{} &
  \multicolumn{1}{c}{Top1$\uparrow$} &
  \multicolumn{1}{c}{Top2$\uparrow$} &
  \multicolumn{1}{c}{Top3$\uparrow$} &
  \multicolumn{1}{c}{} &
  \multicolumn{1}{c}{} \\
  
\midrule
\midrule

\multicolumn{2}{c}{\multirow{10}{*}{HumanML3D}} & MDM \cite{tevet2022human} &ICLR2023         & - & - &0.611$^{\pm.007}$  &0.611$^{\pm.007}$  &5.566$^{\pm.027}$  &9.559$^{\pm.086}$  \\
\multicolumn{2}{c}{}                        & MLD \cite{chen2023executing} &CVPR2023         &0.481$^{\pm.003}$  &0.673$^{\pm.003}$  &0.772$^{\pm.002}$  &0.473$^{\pm.013}$  &3.196$^{\pm.010}$  &9.724$^{\pm.082}$\\
\multicolumn{2}{c}{}                        & ReMoDiffuse \cite{zhang2023remodiffuse} & ICCV2023         &0.510$^{\pm.005}$  &0.698$^{\pm.006}$  &0.795$^{\pm.004}$  &0.103$^{\pm.004}$  &2.974$^{\pm.016}$  &9.018$^{\pm.075}$\\
\multicolumn{2}{c}{}                        & AMD \cite{jingAMDAnatomicalMotion2024} &AAAI2024 & - & - &0.657$^{\pm.006}$  &0.204$^{\pm.001}$  &5.282$^{\pm.032}$  &9.476$^{\pm.077}$\\
\multicolumn{2}{c}{}                        & MotionDiffuse \cite{zhang2024motiondiffuse} &CVPR2024 &0.491$^{\pm.001}$  &0.681$^{\pm.001}$  &0.782$^{\pm.001}$  &0.630$^{\pm.001}$  &3.113$^{\pm.001}$  &9.410$^{\pm.049}$\\
\multicolumn{2}{c}{}                        & EMDM \cite{zhou2025emdm} &ECCV2024 &0.498$^{\pm.007}$  &0.684$^{\pm.006}$  &0.786$^{\pm.006}$  &0.112$^{\pm.019}$  &3.110$^{\pm.027}$  &9.551$^{\pm.078}$\\
\multicolumn{2}{c}{}                        & LADiff \cite{sampieri2025lengthaware} &ECCV2024 &0.493$^{\pm.002}$  &0.686$^{\pm.002}$  &0.784$^{\pm.001}$  &0.110$^{\pm.004}$  &3.077$^{\pm.010}$  &9.622$^{\pm.071}$\\
\multicolumn{2}{c}{}                        & MotionLCM \cite{dai2024motionlcm} &ECCV2024 &0.502$^{\pm.003}$  &0.698$^{\pm.002}$  &0.798$^{\pm.002}$  &0.304$^{\pm.012}$  &3.012$^{\pm.007}$  &9.607$^{\pm.066}$\\
\cmidrule{3-10}
\multicolumn{2}{c}{}                        & Ours (DDPM)& - &0.522$^{\pm.003}$  &\textbf{0.716}$^{\pm.003}$  &\textbf{0.810}$^{\pm.002}$  &0.134$^{\pm.004}$  &\textbf{2.910}$^{\pm.010}$  &9.730$^{\pm.064}$\\
\multicolumn{2}{c}{}                        & Ours & - &\textbf{0.523}$^{\pm.003}$  &0.713$^{\pm.003}$  &0.807$^{\pm.002}$  &\textbf{0.073}$^{\pm.004}$  &2.917$^{\pm.010}$  &\textbf{9.711}$^{\pm.070}$\\
\midrule
\midrule
\multicolumn{2}{c}{\multirow{9}{*}{KIT-ML}} & MDM \cite{tevet2022human} &ICLR2023         & - & - &0.396$^{\pm.004}$  &0.497$^{\pm.021}$  &9.191$^{\pm.022}$  &10.85$^{\pm.109}$\\
\multicolumn{2}{c}{}                        & MLD \cite{chen2023executing} &CVPR2023         &0.390$^{\pm.008}$  &0.609$^{\pm.008}$  &0.734$^{\pm.007}$  &0.404$^{\pm.027}$  &3.204$^{\pm.027}$  &10.80$^{\pm.117}$\\
\multicolumn{2}{c}{}                        & ReMoDiffuse \cite{zhang2023remodiffuse} & ICCV2023         &0.427$^{\pm.014}$  &0.641$^{\pm.004}$  &0.765$^{\pm.055}$  &0.155$^{\pm.006}$  &2.814$^{\pm.012}$  &10.80$^{\pm.105}$\\
\multicolumn{2}{c}{}                        & AMD \cite{jingAMDAnatomicalMotion2024} &AAAI2024 & - & - &0.401$^{\pm.005}$  &0.233$^{\pm.068}$  &9.165$^{\pm.032}$  &10.97$^{\pm.126}$\\
\multicolumn{2}{c}{}                        & MotionDiffuse \cite{zhang2024motiondiffuse} &CVPR2024 &0.417$^{\pm.004}$  &0.621$^{\pm.004}$  &0.739$^{\pm.004}$  &1.954$^{\pm.062}$  &2.958$^{\pm.005}$  &11.10$^{\pm.143}$\\
\multicolumn{2}{c}{}                        & EMDM \cite{zhou2025emdm} & ECCV2024 &0.443$^{\pm.006}$  &0.660$^{\pm.006}$  &0.780$^{\pm.005}$  &0.261$^{\pm.014}$  &2.874$^{\pm.015}$  &10.96$^{\pm.093}$\\
\multicolumn{2}{c}{}                        & LADiff \cite{sampieri2025lengthaware} &ECCV2024 &0.429$^{\pm.007}$  &0.647$^{\pm.004}$  &0.773$^{\pm.004}$  &0.470$^{\pm.016}$  &2.831$^{\pm.020}$  &\textbf{11.30}$^{\pm.108}$\\
\cmidrule{3-10}
\multicolumn{2}{c}{}                        & Ours (DDPM) & - &\textbf{0.462}$^{\pm.007}$  &\textbf{0.682}$^{\pm.006}$  &\textbf{0.800}$^{\pm.005}$  &0.147$^{\pm.008}$  &\textbf{2.625}$^{\pm.017}$  &11.10$^{\pm.094}$\\
\multicolumn{2}{c}{}                        & Ours & - &0.458$^{\pm.006}$  &0.678$^{\pm.006}$  &0.797$^{\pm.005}$  &\textbf{0.122}$^{\pm.004}$  &2.633$^{\pm.017}$  &11.03$^{\pm.099}$\\
\bottomrule
\end{tabular}
}
\end{table*}
\begin{figure*}[t]
  \centering
  \includegraphics[width=\textwidth]{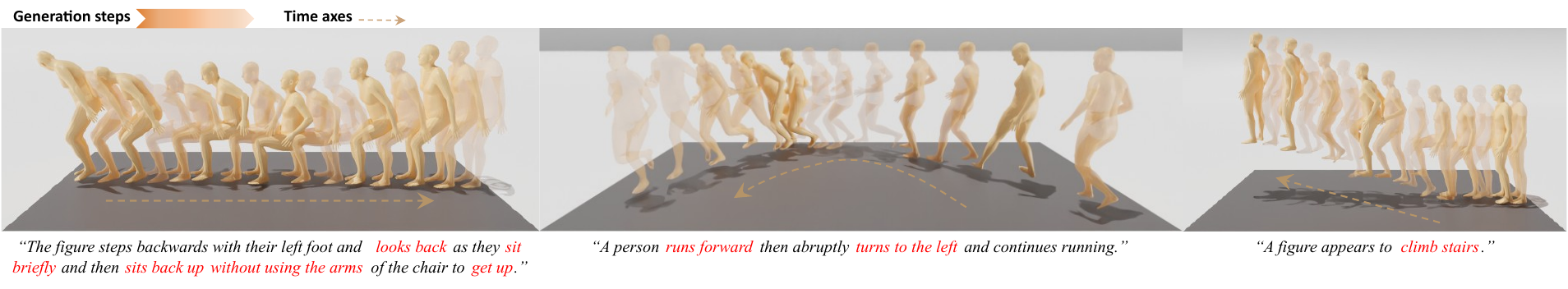}
  \caption{Qualitative results of text-to-motion generation on the HumanML3D dataset.}
  \label{fig:qual_humanml}
  \vspace{-1em}
  \Description{}
\end{figure*}

\subsection{VAE Reconstruction}

To evaluate the reconstruction capability of our continuous motion autoencoder and compare it against existing VQ-VAE architectures, we benchmark several recent state-of-the-art encoders based on different VQ-VAE variants. All models are trained on the HumanML3D dataset, and evaluated on its test set, as well as on the IDEA400 and Kungfu datasets to assess cross-domain generalization. For IDEA400 and Kungfu, we utilize all available data for testing, thereby maximizing the evaluation coverage in unseen domains. 

We report both human skeleton reconstruction metrics and motion generation metrics. with the results summarized in Table~\ref{tab:reconstruction_humanml} and Table~\ref{tab:recons_ood}. Our method demonstrates superior reconstruction performance for both in-domain and out-of-domain actions, exhibiting lower reconstruction error and stronger perceptual alignment. Additional qualitative reconstruction results are provided in the supplementary material.

\subsection{Motion Generation with Keyframe}
Consider a practical scenario: an animator seeks to generate kungfu-style actions using a motion generation model. However, if the model is trained solely on some datasets such as HumanML3D, which do not include kungfu motions, it will likely struggle due to its lack of exposure to such actions. In such case, guiding the model with a small number of reference frames as keyframes with a textual prompt offers an effective solution. To simulate this scenario, we adopt two out-of-distribution motion datasets, \textbf{IDEA400} and \textbf{Kungfu}, both of which share the same body representation as HumanML3D. All models under evaluation are trained exclusively on the HumanML3D training set, with no exposure to the target datasets. During inference, each model is guided with one keyframe per second.

\begin{figure*}[t]
  \centering
  \includegraphics[width=\textwidth]{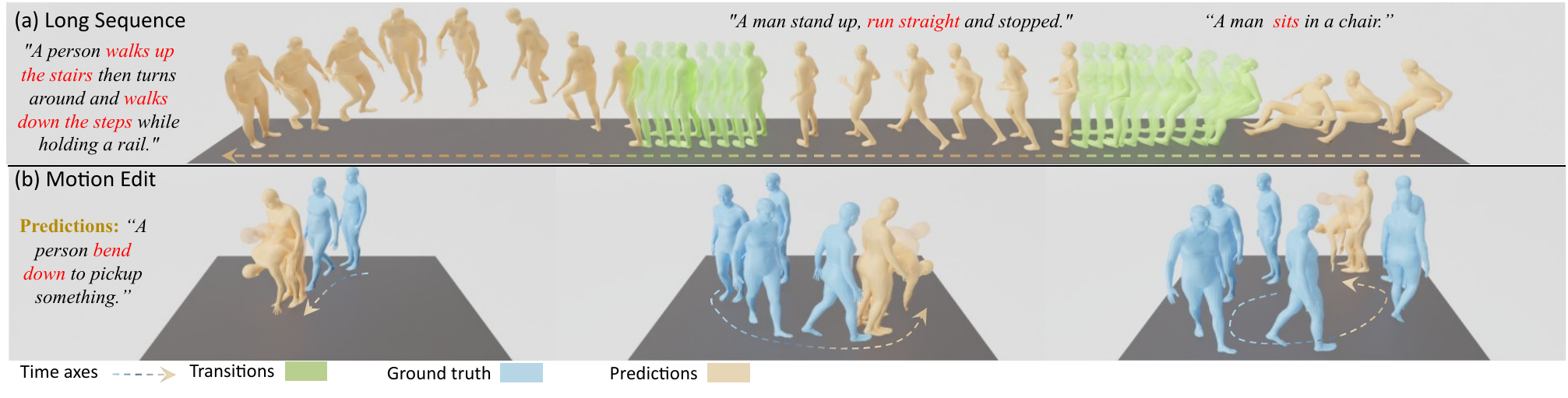}
  \caption{Application examples: (a) long-sequence generation and (b) temporal motion editing with user-specified inputs.}
  \label{fig:applications}
  \Description{}
\end{figure*}


\begin{table}[t]
\caption{Evaluation on the diffusion steps. \textit{T. Steps} indicates training steps. \textit{I. Steps} denotes inference steps.}
\label{tab:diff_steps}
\centering
\resizebox{\columnwidth}{!}{%
\begin{tabular}{lllllll}
\toprule
\multicolumn{1}{c}{T. Steps} &
\multicolumn{1}{c}{I. Steps} &
\multicolumn{1}{c}{$FID$$\downarrow$} &
\multicolumn{1}{c}{Top-3$\uparrow$} &
\multicolumn{1}{c}{MM-Dist$\downarrow$} &
\multicolumn{1}{c}{AITF~(ms)}\\
\midrule
10 & 10 & $0.293{^{\pm.006}}$ & $0.803{^{\pm.002}}$ & $2.971{^{\pm.011}}$ & 1.0682 \\
50 & 50 & $0.134{^{\pm.004}}$ & $0.810{^{\pm.002}}$ & $2.910{^{\pm.010}}$ & 3.2840 \\
100 & 50 & $0.103{^{\pm.004}}$ & $0.806{^{\pm.002}}$ & $2.938{^{\pm.009}}$ & 3.2652 \\
1000 & 50 & $0.108{^{\pm.002}}$ & $0.809{^{\pm.002}}$ & $2.919{^{\pm.002}}$ & 3.2539 \\
100 & 100 & $0.101{^{\pm.004}}$ & $0.805{^{\pm.002}}$ & $2.943{^{\pm.008}}$ & 6.0734 \\
1000 & 100 & $0.099{^{\pm.005}}$ & $0.806{^{\pm.002}}$ & $2.928{^{\pm.008}}$ & 5.8935 \\
\bottomrule
\end{tabular}
\vspace{-1.5em}
}
\end{table}

\begin{figure}[t]
  \centering
  \includegraphics[width=1.0\columnwidth]{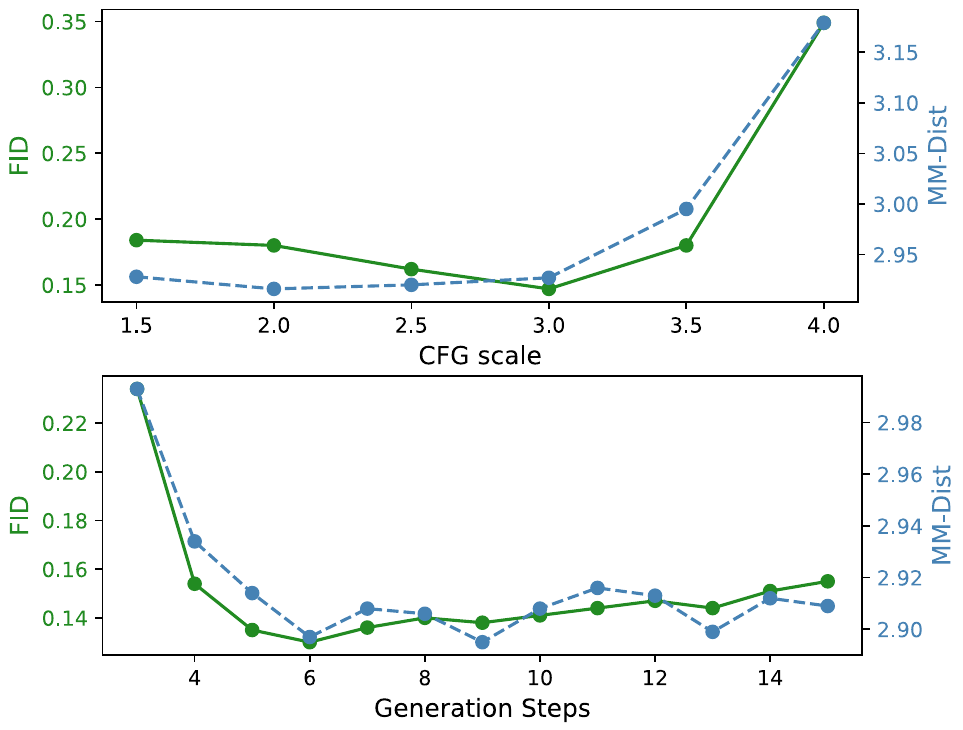}
  \caption{Evaluation of inference parameters, showing the effect of CFG guidance scale $s_c$ and frame generation steps $R$.}
  \label{fig:cfg_steps}
  \Description{}
\end{figure}

For quantitative evaluation, we follow the widely adopted protocol from~\cite{guo2022generating} to train motion-text evaluators for the held-out datasets. The results, presented in Table~\ref{tab:heldout}, demonstrate the generalization ability of each method under sparse keyframe supervision. Our method achieves stronger text-motion alignment, as evidenced by higher R-Precision and lower MM-Dist, while also producing higher-quality motion sequences with lower FID scores. We also provide qualitative results on these two datasets in Figure~\ref{fig:qual_keyframe}. These findings underscore the practical utility of keyframe-based generation, particularly when adapting to novel or out-of-distribution motion domains.

\subsection{Comparison with Existing Methods}

We compare our approach with state-of-the-art diffusion-based motion generation methods that operate either directly on raw motion data~\cite{tevet2022human, zhang2023remodiffuse, han2024amd, zhang2024motiondiffuse, zhou2025emdm, dai2024motionlcm} or on continuous latent representations~\cite{chen2023executing, sampieri2025lengthaware}. Following standard evaluation protocols, we report the results in Table~\ref{tab:standard_bench}. Overall, our method consistently outperforms existing approaches across several key metrics. In particular, it achieves superior text-motion alignment (lower MM-Dist), higher motion quality (lower FID), and stronger semantic consistency with input text (higher R-Precision), demonstrating its effectiveness in generating coherent and high-quality motions. We illustrate some motions generated with our method in Figure~\ref{fig:qual_humanml}. For more qualitative results, please refer to the supplementary.

\subsection{Ablation Studies}
To analyze the contribution of individual components and design choices, we conduct ablation studies on the HumanML3D evaluation protocol.

\subsubsection{Inference Parameters} During inference, two key hyperparameters influence performance: the classifier-free guidance (CFG) scale factor $s_c$ and the number of frame generation steps $R$. We assess their impact on the HumanML3D test set using FID and MM-Dist scores, as illustrated in Figure~\ref{fig:cfg_steps}. The results show that performance peaks at approximately $s_c = 3.0$; deviations from this value in either direction lead to a decline in generation quality. Regarding $R$, increasing the number of generation steps improves performance up to $R=10$, beyond which the benefit diminishes. Our next-batch generation strategy helps reduce the number of required autoregressive steps, thereby enhancing inference efficiency without sacrificing quality.

\subsubsection{Diffusion Steps} The number of diffusion steps $T$ is a key factor that balances motion fidelity and inference speed. We train our model with different values of $T$ and inference with both DDPM and DDIM sampling strategies. Table~\ref{tab:diff_steps} reports FID, MM-Dist, Top-3 Accuracy, and Average Inference Time per Frame (AITF) on the HumanML3D test set. The results indicate that increasing $T$ during training generally enhances performance, albeit with longer inference times. To alleviate this issue, we employ DDIM sampling at test time, which significantly accelerates inference while maintaining competitive generation quality.

\subsection{More Applications}

Our model enables fine-grained temporal control and can be extended to various applications. In this section, we demonstrate several use cases of our approach.

\subsubsection{Long Motion Generation} Our method supports generating motions of arbitrary length through a generate \& stitch paradigm, as illustrated in Figure~\ref{fig:applications}(a). In the first stage, the model generates motion clips based on different text prompts. In the second stage, it stitches adjacent clips by generating smooth transition frames using the last few frames of the preceding clip and the first few frames of the following one.

\subsubsection{Temporal Motion Editing} Due to the strong representation capability of continuous VAE, our model allows users to specify partial motion sequences and edit or extend them accordingly. As shown in Figure~\ref{fig:applications}(b), users can define the number of ground-truth frames to preserve, and the model will seamlessly generate the remaining motion to complete the sequence.

\section{Conclusion}

In this paper, we propose MoMADiff, a Motion Masked Autoregressive Diffusion model for text-guided human motion generation. MoMADiff generates frame-level continuous motion representations, allowing fine-grained spatial and temporal control of synthesized motions. Our model demonstrates strong robustness on out-of-distribution motions, maintaining high controllability with respect to user-defined text and motion prompts. These capabilities enable a wide range of applications, including keyframe-based motion generation, long-form motion synthesis, and temporal motion editing.

\bibliographystyle{ACM-Reference-Format}
\balance
\bibliography{acmart}

\clearpage
\appendix
\section*{Appendix}
\addcontentsline{toc}{section}{Appendix}

\begin{center}
  {\LARGE \textbf{Supplementary Material}}\\[1em]
\end{center}

\renewcommand\thesection{\Alph{section}} 
\setcounter{section}{0}
\setcounter{equation}{1}

\setcounter{figure}{0}
\setcounter{table}{0}
\renewcommand{\thefigure}{\Roman{figure}}
\renewcommand{\thetable}{\Roman{table}}

\section{More Ablation Studies}

\subsection{Key Components}

The components of MoMADiff are intentionally designed to be closely integrated. To evaluate the contribution of each component, we performed an ablation study by selectively removing or replacing them. Specifically, we trained the following three variants:

\textbf{Without VAE}: We removed the VAE component, which is responsible for modeling the spatial structure and local motion priors. In this setting, the model directly predicts the motion sequence $X$ in the observation space, instead of modeling its latent representation.

\textbf{Without Diffusion Head}: We removed the diffusion head, so the Transformer directly models the latent variable $Z$ without first producing condition vectors for diffusion.

\textbf{Transformer Only}: We removed both the VAE and diffusion head, leaving only a Transformer. In this case, the model takes masked motion sequences as input and recursively predicts the motion $X$ in an autoregressive manner.

The results are presented in Table~\ref{tab:vae_diffusion_ablation}. We observe that the inclusion of the VAE provides useful motion priors, leading to improved performance compared to the baseline without the VAE and diffusion head. However, it remains challenging for the Transformer to predict directly in the continuous motion space. By introducing the diffusion head to handle the generation of continuous representations, the Transformer can instead focus on modeling temporal dependencies. The diffusion process produces more accurate and reliable motion representations, leading to better generation results.

\subsection{Inference Modes}

\begin{figure}[th]
  \centering
  \includegraphics[width=\linewidth]{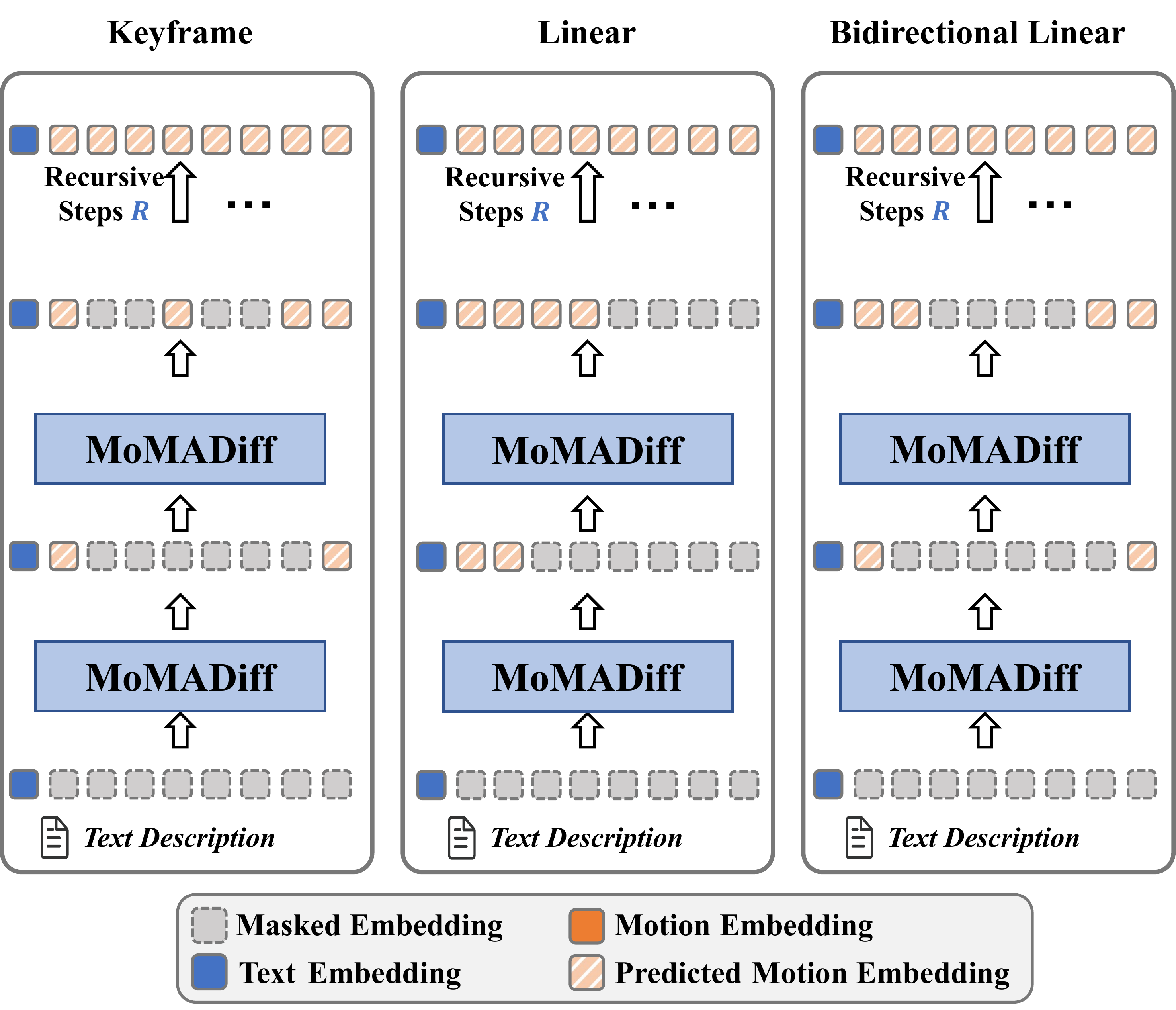}
  \caption{Illustration of different types of inference modes.}
  \label{fig:infer_mode}
  \Description{}
\end{figure}

We use a cosine scheduler to determine the number of frames to be predicted at each iteration, selecting them randomly from the motion sequence. Additionally, we explore several alternative inference modes to better understand how different generation orders affect performance, as illustrated in Figure~\ref{fig:infer_mode}.

\textbf{Keyframe Mode.} In this mode, the masking ratio at the $i$-th step is determined by the function $y=cos(\frac{\pi}{2}\cdot\frac{i}{R})$, where $R$ is the total number of steps. The motion embeddings to be predicted are randomly selected from the sequence, simulating a sparse keyframe first generation process.

\textbf{Linear Mode.} This mode uses a linear function $y=1-\frac{i}{R}$ to control the number of masked frames at each step. The motion embeddings to be predicted are selected sequentially from the beginning to the end of the sequence, following a next-set generation strategy.

\textbf{Bidirectional Linear.} Similar to the Linear mode, this approach also uses $y=1-\frac{i}{R}$ to schedule the masking ratio. However, instead of predicting frames in a single direction, the motion embeddings are selected symmetrically from both the beginning and end of the sequence. Generation progresses inward from both sides in a bidirectional next-set manner.

Quantitative results are presented in Table~\ref{tab:inference_mode}. As shown, keyframe mode consistently demonstrates superior performance compared to the other inference strategies.

\subsection{Depth of the Diffusion Head}

We conduct ablation studies on the number of layers in the diffusion head, with results summarized in Table~\ref{tab:diffusion_head}. All experiments are based on the DDPM model using 50 diffusion steps. We observe that the best performance is achieved with four layers of diffusion blocks. Using fewer blocks may limit the model's capacity to capture motion dynamics, while increasing the number of layers can lead to gradient vanishing issues, likely due to our use of a simple MLP-based design without skip connections.

Additionally, we experimented with the diffusion head architecture proposed in~\cite{li2024autoregressive}, originally designed for image generation. However, its skip connection structure did not yield satisfactory results in our motion generation task, suggesting that architectural designs optimized for image domains may not transfer well to motion modeling.

\section{Inference Speed}
\begin{table*}[tbh]
\caption{Ablation study on the impact of VAE and Diffusion modules}
\label{tab:vae_diffusion_ablation}
\centering
\begin{tabular}{cc|ccc|ccc}
\toprule
VAE & Diffusion & Top-1$\uparrow$ & Top-2$\uparrow$ & Top-3$\uparrow$ & FID$\downarrow$ & MM-Dist$\downarrow$ & Diversity$\uparrow$ \\
\midrule
\checkmark & \checkmark & \textbf{0.523}$\pm$.003 & \textbf{0.713}$\pm$.003 & \textbf{0.807}$\pm$.002 & \textbf{0.073}$\pm$.004 & \textbf{2.917}$\pm$.010 & \textbf{9.711}$\pm$.070 \\
  & \checkmark & 0.397$\pm$.003 & 0.571$\pm$.003 & 0.673$\pm$.002 & 1.171$\pm$.014 & 3.864$\pm$.010 & 9.181$\pm$.082 \\
\checkmark &   & 0.440$\pm$.003 & 0.631$\pm$.003 & 0.736$\pm$.002 & 0.949$\pm$.017 & 3.469$\pm$.008 & 9.616$\pm$.079 \\
  &   & 0.390$\pm$.003 & 0.570$\pm$.003 & 0.679$\pm$.003 & 1.483$\pm$.014 & 3.789$\pm$.012 & 9.511$\pm$.087 \\
\bottomrule
\end{tabular}
\end{table*}

MoMADiff supports flexible trade-offs between efficiency and quality by adjusting two factors: the number of recursive steps (R) and the number of DDIM inference steps (I. Steps). To quantify computational cost, we report the Average Inference Time per Sentence (AITS) in Table~\ref{tab:baselines}.

As shown, increasing either R or I. Steps improve generation quality (e.g., lower FID) but also increase inference time, providing users with the flexibility to balance performance and speed according to practical needs. Under a fast inference setting (I. Steps = 10, R=3), our model already achieves slightly better FID scores than previous baselines with comparable inference time. Furthermore, with more DDIM and recursive steps (e.g., I. Steps = 10, R = 9), our model continues to improve in quality while maintaining competitive efficiency. This design also supports fine-grained control over the generation process, allowing dynamic adjustments between speed and quality.

\begin{table}[!ht]
\caption{Comparison with prior methods on motion generation}
\label{tab:baselines}
\centering
\begin{tabular}{lcc}
\toprule
Method & FID$\downarrow$ & AITS$\downarrow$ \\
\midrule
TEMOS                    & 3.734 & 0.017 \\
T2M                      & 1.067 & 0.038 \\
MotionDiffuse            & 0.630 & 14.740 \\
MDM                      & 0.544 & 24.740 \\
MLD                      & 0.473 & 0.217 \\
MotionLCM                & 0.467 & 0.030 \\
Ours (I. Steps 10, R=3)  & 0.329 & 0.074 \\
Ours (I. Steps 50, R=9)  & 0.107 & 0.778 \\
Ours (I. Steps 100, R=9) & \textbf{0.073} & \textbf{1.483} \\
\bottomrule
\end{tabular}
\end{table}

  


\begin{table*}[tbh]
\caption{Ablation on Inference Mode.}
\label{tab:inference_mode}
\centering
\begin{tabular}{lcccccc}
\toprule
\multirow{2}{*}{Methods} & \multicolumn{3}{c}{R-Precision$\uparrow$} & \multirow{2}{*}{FID$\downarrow$} & \multirow{2}{*}{MM-Dist$\downarrow$} & \multirow{2}{*}{Diversity$\uparrow$} \\
\cmidrule(lr){2-4}
& Top1 & Top2 & Top3 &  &  &  \\
\midrule
Keyframe & \textbf{0.523}$^{\pm.003}$  &\textbf{0.713}$^{\pm.003}$  &\textbf{0.807}$^{\pm.002}$  &\textbf{0.073}$^{\pm.004}$  &\textbf{2.917}$^{\pm.010}$  &\textbf{9.711}$^{\pm.070}$ \\
Linear & 0.514$^{\pm.004}$ & 0.707$^{\pm.003}$ & 0.804$^{\pm.002}$ & 0.131$^{\pm.003}$ & 2.938$^{\pm.009}$ & 9.555$^{\pm.067}$ \\
Bi-directional Linear & 0.512$^{\pm.003}$ & 0.703$^{\pm.003}$ & 0.800$^{\pm.002}$ & 0.108$^{\pm.005}$ & 2.960$^{\pm.009}$ & 9.580$^{\pm.069}$ \\
\bottomrule
\end{tabular}

\end{table*}

\begin{figure*}[t]
  \centering
  \includegraphics[width=0.9\textwidth]{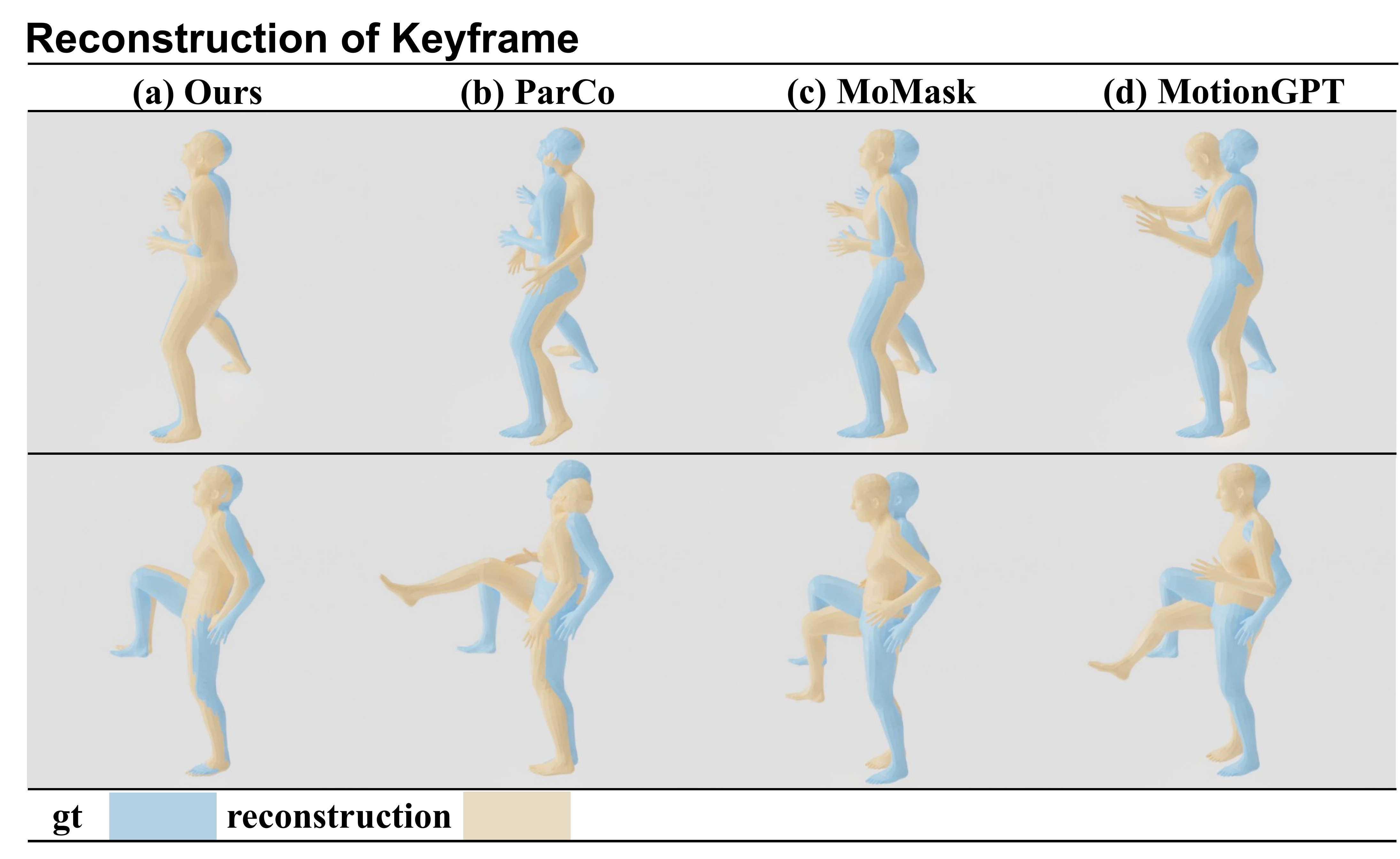}
  \caption{Additional motion reconstruction results on out-of-distribution motions using different encoders.}
  \label{fig:reconst}
  \Description{}
\end{figure*}

\begin{table*}[tbhp]
\caption{Ablation on diffusion head design}
\label{tab:diffusion_head}
\centering
\begin{tabular}{lcccccc}
\toprule
\multirow{2}{*}{Setting} & \multicolumn{3}{c}{R-Precision$\uparrow$} & \multirow{2}{*}{FID$\downarrow$} & \multirow{2}{*}{MM-Dist$\downarrow$} & \multirow{2}{*}{Diversity$\uparrow$} \\
\cmidrule(lr){2-4}
& Top1 & Top2 & Top3 &  &  &  \\
\midrule
Ours with 3 layers  & 0.497$^{\pm.003}$  &0.689$^{\pm.002}$  &0.785$^{\pm.002}$  &0.266$^{\pm.008}$  &3.148$^{\pm.011}$  &\textbf{9.819}$^{\pm.073}$ \\
Ours with 4 layers & \textbf{0.522}$^{\pm.003}$  &\textbf{0.716}$^{\pm.003}$  &\textbf{0.810}$^{\pm.002}$  &\textbf{0.134}$^{\pm.004}$  &\textbf{2.910}$^{\pm.010}$  &9.730$^{\pm.064}$ \\
Ours with 6 layers & 0.516$^{\pm.003}$ & 0.708$^{\pm.002}$ & 0.802$^{\pm.002}$ & 0.191$^{\pm.005}$ & 2.945$^{\pm.008}$ & 9.737$^{\pm.079}$ \\
Diffusion head in~\cite{li2024autoregressive} & 0.450$^{\pm.003}$ & 0.633$^{\pm.004}$ & 0.735$^{\pm.003}$ & 0.675$^{\pm.017}$ & 3.361$^{\pm.010}$ & 9.114$^{\pm.070}$ \\
\bottomrule
\end{tabular}

\end{table*}

\section{Additional Qualitative Results}

\subsection{Ours on HumanML3D}

\begin{figure*}[!t]
  \centering
  \includegraphics[width=\textwidth]{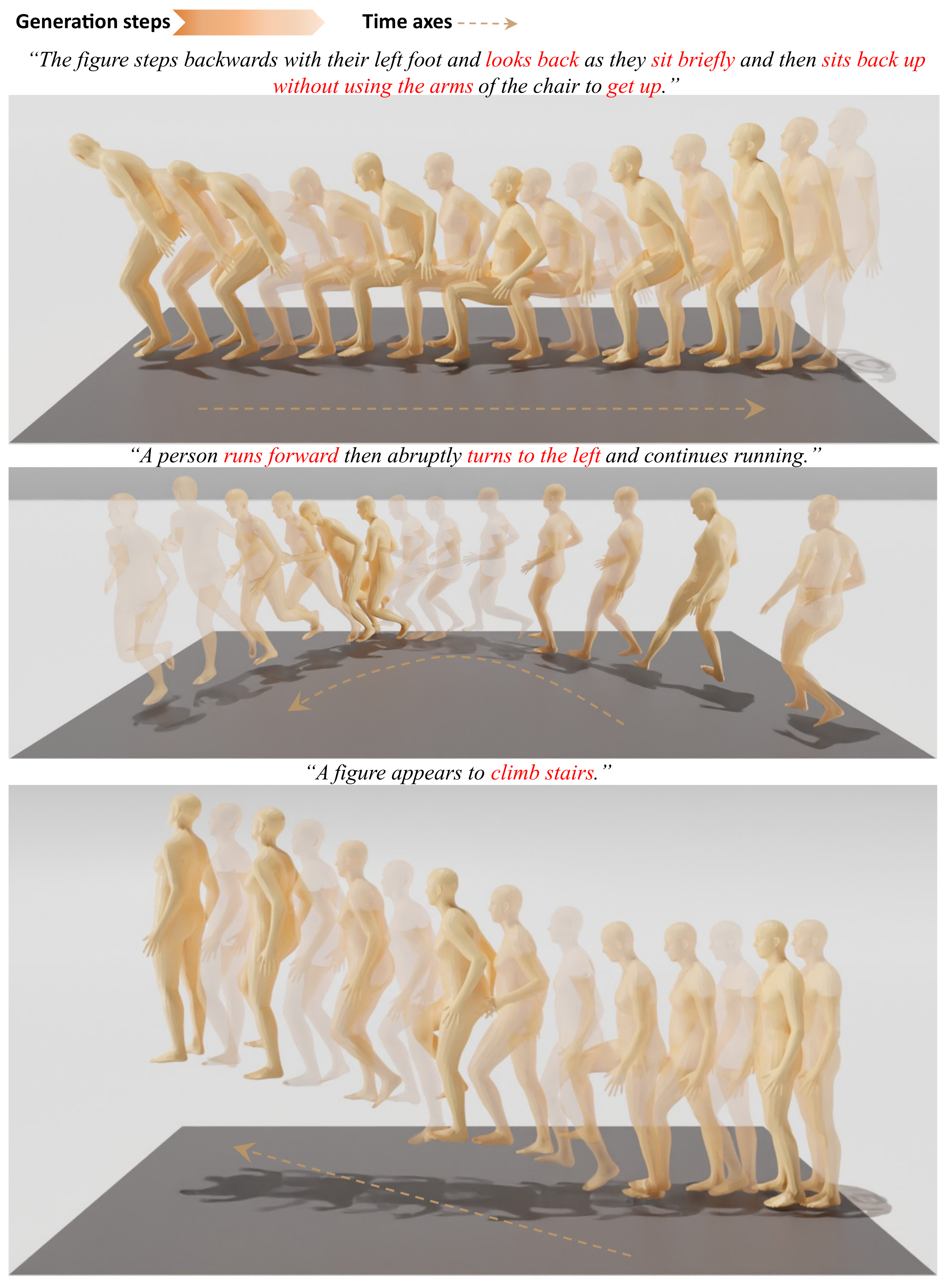}
  \caption{Qualitative results of text-to-motion generation on the HumanML3D dataset.}
  \label{fig:qual_humanml_large}
  \Description{}
\end{figure*}

We provide qualitative results of our method on HumanML3D dataset in Figure~\ref{fig:qual_humanml_large}. Please also visit our project page for more qualitative video results, which include comparisons of motion reconstruction, keyframe-guided generation, and text-to-motion generation on the HumanML3D dataset.

\subsection{Reconstruction Results}

We further illustrate the reconstruction capability of our continuous motion autoencoder by comparing it against existing VQ-VAE architectures with two samples on the Kungfu dataset. As shown in Figure~\ref{fig:reconst}, our method achieves closer reconstructions to the ground truth compared to previous encoder-based approaches, demonstrating its superior ability to preserve details of out-of-distribution motions.

\section{More Implementation Details}
\label{sec:PGTA method}

\subsection{Evaluation Metrics}

We adopt the motion-text evaluator from~\cite{guo2022generating} and use the following evaluation metrics.

(1) \textit{R-Precision} Measures text-motion alignment by computing Euclidean distances between a motion feature and 32 candidate text features. We report top-1, top-2, and top-3 retrieval accuracies.

(2) \textit{Frechet Inception Distance (FID)}~\cite{heusel2017gans}: Evaluates the distributional similarity between generated and real motions, based on features extracted by the motion encoder.

(3) \textit{Multimodal Distance (MMD):} Calculates the average Euclidean distance between motion features and their corresponding text features.

(4) \textit{Diversity:} Assesses the variety in generated motions by computing the average Euclidean distance between 300 randomly sampled pairs.

\subsection{Datasets}

\textbf{HumanML3D} dataset collects the motions from AMASS~\cite{mahmood2019amass} dataset and HumanAct12~\cite{guo2020actionmotion} dataset, which contains 14,616 human motions. The dataset provides 44,970 textual descriptions in total for these motions, with three descriptions for each motion sequence. HumanML3D contains diverse actions including daily activities, sports, acrobatics, and artistry. 

\textbf{KIT-ML} dataset contains 3,911 motions from KIT~\cite{mandery2015kit} and CMU~\cite{carnegie}, and includes 6,278 descriptions. 

\textbf{IDEA400} dataset is the largest subset apart from AMASS dataset, which contains 12,042 motion sequences with one text description each. It contains 400 actions with human self-contact motions and human-object contact motions during walking, standing, and sitting, which examine the detailed modeling capability of the model. 

\textbf{Kungfu} includes 1,032 motion sequences with one semantic text description per sequence. It represents a highly challenging out-of-distribution evaluation scenario due to its complex and stylized motion patterns.

IDEA400 and Kungfu are derived from the high-quality, whole-body, large-scale human motion dataset Motion-X. To ensure compatibility with HumanML3D, we extract body-only poses and semantic text descriptions and format the data to match the HumanML3D specification.

Since motions in the HumanML3D dataset have a maximum length of 196 frames, we follow this constraint when evaluating on the IDEA400 and Kungfu datasets by selecting only motion sequences shorter than 196 frames.

\begin{figure*}[ht]
\centering
\begin{minipage}{\textwidth}
\begin{Verbatim}[frame=single]
Hardware Environment
--------------------
CPU: Intel(R) Xeon(R) Gold 5218R CPU @ 2.10GHz
Memory: 256 GB
GPU: GeForce RTX 3090

Software Environment
--------------------
sys.platform: linux
Python: 3.10.14 [GCC 11.2.0]
CUDA available: True
GPU 0,1: NVIDIA GeForce RTX 3090
GCC: gcc (Ubuntu 7.5.0-3ubuntu1~18.04) 7.5.0
PyTorch: 2.4.0+cu121
PyTorch compiling details: PyTorch built with:
- GCC 9.3
- C++ Version: 201703
- Intel(R) oneAPI Math Kernel Library Version 2022.2-Product Build 20220804 for Intel(R) 64 
architecture applications
- Intel(R) MKL-DNN v3.4.2
- OpenMP 201511 (a.k.a. OpenMP 4.5)
- LAPACK is enabled (usually provided by MKL)
- NNPACK is enabled
- CPU capability usage: AVX512
- CUDA Runtime 12.1
- NVCC architecture flags: -gencode;arch=compute_50,code=sm_50;-gencode;arch=compute_60,
code=sm_60;-gencode;arch=compute_70,code=sm_70;-gencode;arch=compute_75,code=sm_75;-gencode;
arch=compute_80,code=sm_80;-gencode;arch=compute_86,code=sm_86;-gencode;arch=compute_90,
code=sm_90
- CuDNN 90.1  (built against CUDA 12.4)
- Magma 2.6.1

TorchVision: 0.19.0+cu121
\end{Verbatim}
\caption{Hardware and software environments.}
\label{fig:envir}
\end{minipage}
\end{figure*}

\subsection{Hardware and Software Environments}

The hardware and software environments used in our experiments are illustrated in Figure~\ref{fig:envir}. All processes, including training and inference, are conducted on machines with these configurations.

\end{document}